\pdfoutput=1
\documentclass[letterpaper, 10pt, conference]{ieeeconf}

\usepackage[space,noadjust]{cite}
\usepackage{color}
\usepackage[usenames,dvipsnames]{xcolor}
\usepackage[linkcolor=RoyalPurple,citecolor=black,urlcolor=black,colorlinks=true]{hyperref}

\usepackage{algorithm}
\usepackage{algpseudocode}
\usepackage{amsmath}
\usepackage{amssymb}

\usepackage{subcaption}
\usepackage{graphicx}
\usepackage{graphics}
\usepackage{float}

\usepackage{multirow}

\IEEEoverridecommandlockouts

\newcommand{\etal}{\textit{~et al.}}

\def\real{\mathbb{R}}

\newcommand{\norm}[1]{\left\lVert#1\right\rVert}

\title{\LARGE \bf 
Decentralized Global Connectivity Maintenance for Multi-Robot Navigation: A Reinforcement Learning Approach}
\author{Minghao Li, Yingrui Jie, Yang Kong, Hui Cheng$^{\star}$
\thanks{The authors are with the School of Computer Science and Engineering and the School of Eletronics and Communication Engineering, Sun Yat-sen University, Guangzhou and Shenzhen, China}
\thanks{$^{\star}$Corresponding to: \tt\small{chengh9@mail.sysu.edu.cn}}
}

\begin{document}

\maketitle

\begin{abstract}
The problem of multi-robot navigation of connectivity maintenance 
is challenging in multi-robot applications.
This work investigates how to navigate a multi-robot
team in unknown environments while maintaining connectivity.
We propose a reinforcement learning (RL) approach to develop a
decentralized policy, which is shared among multiple robots. 
Given range sensor measurements and the positions of other robots, 
the policy aims to generate control commands for navigation and preserve the global connectivity of the robot team.
We incorporate connectivity concerns into the RL framework as constraints and introduce behavior cloning to reduce the exploration complexity of policy optimization.
The policy is optimized with all transition data collected by multiple robots in random simulated scenarios.
We validate the effectiveness of the proposed approach by comparing different combinations of connectivity constraints and behavior cloning. We also show that our policy can generalize to unseen scenarios in both simulation and holonomic robots experiments.

\end{abstract}


%
\IEEEpeerreviewmaketitle

\section{Introduction}

Multi-robot navigation is one of the significant problems for multi-robot systems and
practical to cooperative tasks such as search and rescue, formation control, and area coverage. As each robot has limited communication and sensing ranges, 
cooperative behaviors among robots rely on the connectivity of the communication network.
Therefore, maintaining connectivity over time is important for multi-robot navigation.

The problem of connectivity maintenance for multi-robot systems has been widely studied, and various connectivity 
strategies~\cite{li2013bounded,sabattini2013decentralized,williams2015global,gasparri2017bounded} have been proposed. 
Most of these works consider global connectivity because of its flexibility as 
it allows robots to break communication links
as long as the communication graph is connected over time.
This is guaranteed by preserving the second-smallest eigenvalue of the underlying graph Laplacian of robot networks to be positive, which is well-known as algebraic connectivity~\cite{fiedler1973algebraic}. 
As these works attempt to complete various tasks,
some of them validate their strategies in obstacle-free environments for formation control~\cite{gasparri2017bounded}, visiting multiple waypoints~\cite{wang2016multi} and area coverage~\cite{capelli2020connectivity}.
Potential field methods are widely studied to combine connectivity maintenance with system objectives and obstacle avoidance.
Williams\etal~\cite{williams2015global} select a dispersive potential with collision avoidance to validate the designed connectivity controller at the presence of disc-obstacles. Li\etal~\cite{li2013bounded} have proposed a potential-based method towards circle obstacles avoidance with connectivity maintenance.
Sabattini\etal~\cite{sabattini2013decentralized} evaluate their
gradient-based controller in rendezvous applications.
They use a collision avoidance strategy which is not mentioned in experiments as the controller is designed for general purposes.
\begin{figure}[t]
    \centering
    \includegraphics[width=0.4\textwidth]{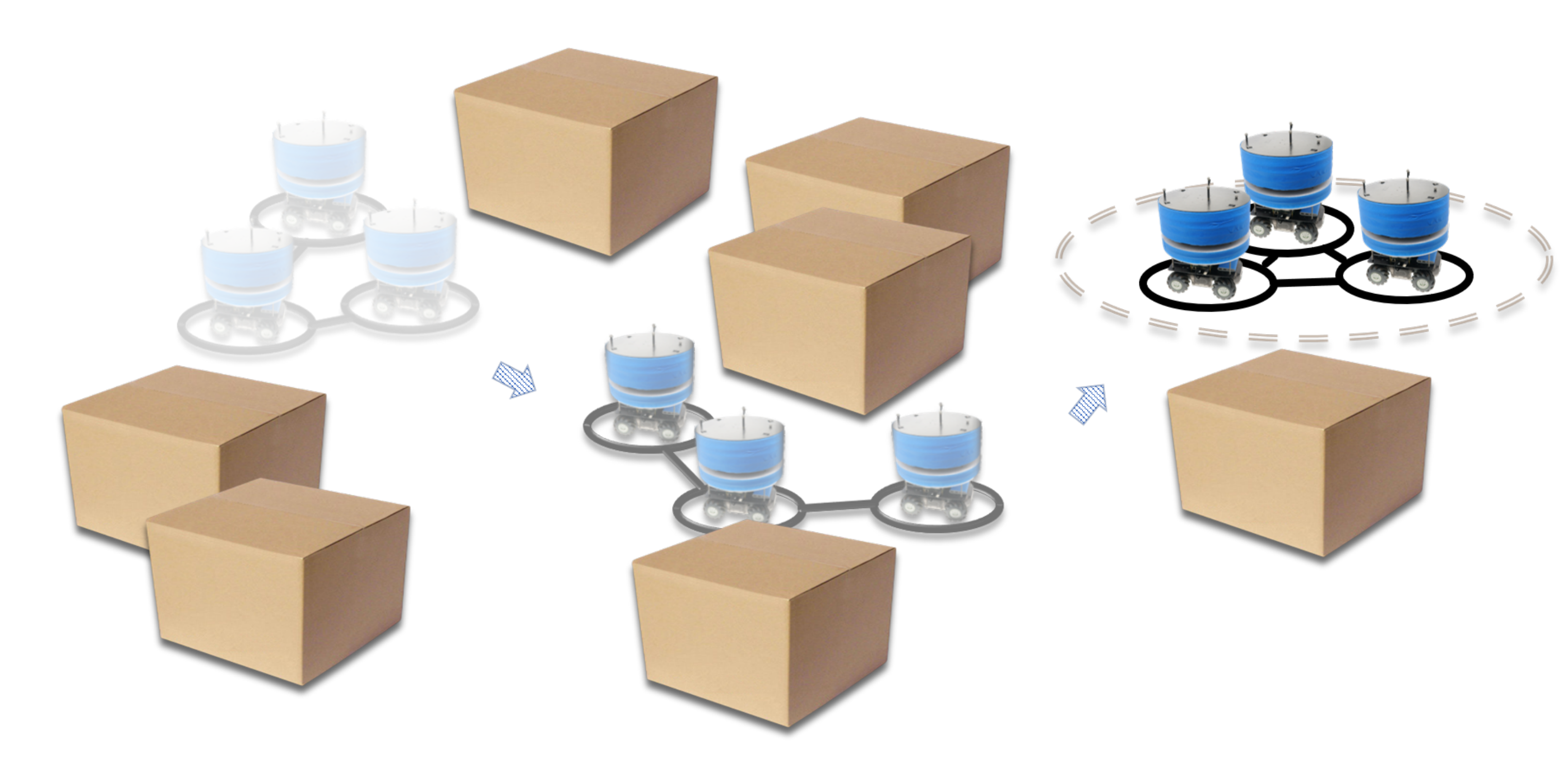}
    \caption{Illustration of our research task. The proposed decentralized policy navigates multiple robots to a destination with global network connectivity maintenance.}
    \label{cover}
    \vspace{-7mm}
\end{figure}
One of the major challenges of multi-robot navigation is the imperfect environment
information due to limited sensing capabilities. Connectivity controllers may fail since they can be trapped at the presence of obstacles without the explicit design of navigation strategies. Moreover, simple assumptions on obstacle shapes have limitations in complex environments. 

To navigate in complex environments, 
RL-based approaches are alternatives that
have shown promising results in map-less navigation. 
Many of these works have developed end-to-end navigation policies which
directly map raw sensor measurements to control commands without prior knowledge
of the environment~\cite{pfeiffer2018reinforced, long2018towards,lin2019end}.
While few of them focus on connectivity of the robot team, Lin\etal~\cite{lin2020connectivity} propose a RL framework that combines a constrained optimization module to enforce local connectivity constraints. However, local connectivity may lead to restrictive behaviors since each
robot has to maintain communication links with all other robots in the team over time. Moreover, the effects of action correction from the optimization module are nontrivial in RL as it may break the Markov assumption~\cite{cheng2019end, mysore2020regularizing}.

In this work, we focus on multi-robot navigation with connectivity maintenance 
at the presence of unknown obstacles.
We consider global connectivity to allow flexible changes of communication graphs
and incorporate it as constraints into RL methods. 
Although the connectivity-constrained multi-robot navigation problem can be formulated as a Constrained Partial Observable Markov Decision Process (Constrained POMDP), existing constrained RL methods are infeasible due to sample inefficiency and the inherent difficulty of this multi-objective problem (reaching target points and avoiding collisions while bounding connectivity constraints).
To address this issue, we present a model-free constrained RL approach
that combines behavior cloning (BC) to make policy training tractable.
By learning from demonstrations, the actions of robots are guided towards expert navigation behaviors and therefore improve efficiency in policy search.
Following the centralized training decentralized execution  scheme~\cite{lowe2017multi,foerster2016learning}, a shared policy is optimized by experiences of all agents during training.
As a result, a decentralized policy is developed that takes information of other agents into account to maintain global connectivity and leverages range sensor data for collision avoidance.

The contributions of this work are:
\begin{itemize}
    \item A constrained RL-based approach is proposed to 
    navigate multiple robots in unknown environments while maintaining global connectivity in a decentralized manner.
    \item An effective way to facilitate policy training by combing BC and RL
    without breaking the connectivity constraint.
    \item The performance of the developed policy is empirically evaluated in simulation and real-world robotic platforms.
\end{itemize}


\section{Related Work}
RL-based approaches have been intensively studied in the last few years and have shown great potentials in navigation with imperfect sensing in unknown complex environments.
For single-robot navigation,
Pfeiffer\etal~\cite{pfeiffer2018reinforced} propose a sample efficient approach
that uses expert demonstrations for policy pre-training and further improve the
policy performance by RL. The experiments show that the policy outperforms their previous approach, where an end-to-end policy is trained by supervised learning only~\cite{pfeiffer2017perception}. In particular, they treat collision avoidance as constraint to ensure safety and use pre-training to improve sample efficiency. In our approach, a connectivity-maintained expert is infeasible, and thus we impose connectivity constraints and provide navigation demonstrations in one-stage RL training without pre-training.
Zhang\etal~\cite{zhang2020map} aim to develop a flexible policy and propose a transfer method that can handle various range sensor dimensional configurations. Later, they propose a support-point approach to distinguish critical points from LiDAR range measurements and handle different LiDAR configurations without re-training.

In multi-robot settings,
Chen\etal~\cite{chen2017decentralized} propose an approach called CADRL that considers observable states of other robots and pedestrians instead of direct mapping from perception to motion commands. With a trained value function, the robots perform one-step lookahead to generate collision-free actions. Later, they extend CADRL and design a socially-aware reward function for learning a policy that incorporates social norms towards pedestrian-rich navigation. 
Everett\etal~\cite{everett2018motion} augment this framework with Long short-term memory to improve performance on complex interactions with pedestrians.
The parameter sharing scheme~\cite{gupta2017cooperative,nguyen2020deep} is popular for efficient multi-agent policy training and applicable to multi-robot navigation.
Long\etal~\cite{long2018towards} train a multi-robot collision avoidance policy with a multi-agent extended variant of Proximal Policy Optimization (PPO) and adopt curriculum learning to facilitate training. 
This policy is shared among robots and navigates each robot to its desired goal 
from 2D laser measurements without communications in a decentralized manner.
Lin\etal~\cite{lin2019end} aim to navigate the geometric center of a robot team to reach waypoints and develop an end-to-end policy shared among the robots that consider raw laser data inputs and position information of other robots.  Han\etal~\cite{han2020cooperative} also consider observable states of other robots and dynamic obstacles with Gaussian noises. They propose a greedy target allocation method to efficiently navigate the robots to multiple targets.
\vspace{-0.5mm}
\section{Approach}
\begin{figure}[t]
    \centering
    \includegraphics[width=0.43\textwidth]{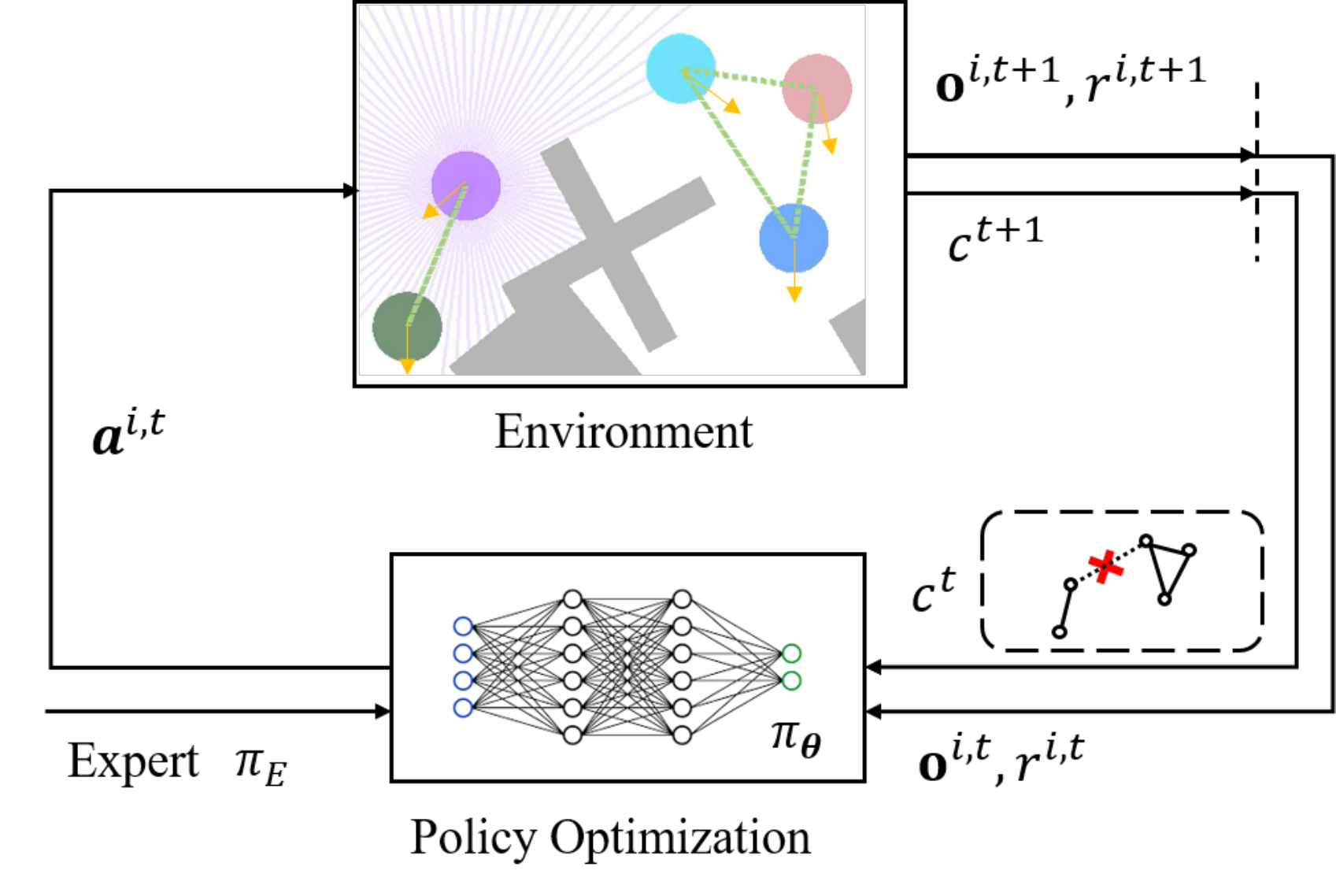}
    \vspace{-2mm}
    \caption{An overview of our approach. Each robot computes its control command $\mathbf{a}^{i,t}$ with the shared policy $\pi_{\boldsymbol\theta}$ given its observation $\mathbf{o}^{i,t}$ and receives reward $r^{i,t+1}$ and connectivity signal $c^{t+1}$. The policy $\pi_{\boldsymbol\theta}$ is optimized by the experiences of all robots under supervised actions of an expert $\pi_E$ and connectivity constraints. }
    \vspace{-6.5mm}
    \label{fig:overview} 
\end{figure}

\subsection{Problem formulation}
The multi-robot navigation problem can be formulated as  
a Partial Observable Markov Decision Process (POMDP).
Formally, a POMDP is defined by tuple 
$<\mathcal{S}, \mathcal{A}, \mathcal{O}, \mathcal{P}, \mathcal{Z}, r, \gamma>$
where $\mathcal{S}$ is the state space, $\mathcal{A}$ is the action space,
$\mathcal{O}$ is the observation space, 
$\mathcal{P}(s^\prime|s,a)$ is the transition probabilities of taking action
$a$ at state $s$, 
$\mathcal{Z}(o|s^\prime,a)$ is the observation distribution at the new state $s^\prime$, 
$r(s,a):\mathcal{S}\times\mathcal{A}\to \real$ is the reward function 
and $\gamma\in(0,1]$ is the discount factor.

Specifically, we consider a team of $N$ robots navigate to a target zone $\mathbf{g}$ with radius $R_g$. The robots are considered homogeneous and modeled as discs with the same radius $R_r$ and communication range $R_c$.
At each timestep $t$, 
each robot $i$ obtains 2D laser scans $\mathbf{o}_l^{i,t}$ 
and its velocity $\mathbf{o}_v^{i,t}$.
The relative goal position $\mathbf{o}_g^{i,t}$
and the positions of other robots in the team $\mathbf{o}_p^{i,t} = [\mathbf{p}_1^t, \dots,\mathbf{p}_{i-1}^t,\mathbf{p}_{i+1}^t,\dots,\mathbf{p}_N^t]$
are informed to the robot $i$. 
Given the observation $\mathbf{o}^{i,t} = [\mathbf{o}_l^{i,t}, \mathbf{o}_v^{i,t}, \mathbf{o}_g^{i,t}, \mathbf{o}_p^{i,t}]$, the robot aims to obtain a parameterized policy $\pi_{\boldsymbol\theta}$ that directly maps $\mathbf{o}^{i,t}$ to control command $\mathbf{a}^{i,t}$.
The policy $\pi_{\boldsymbol\theta}$ is shared among robots, and each robot is allowed to take different actions since they receive different observations.

The objective of RL is to find the optimal parameter $\boldsymbol\theta$, where the corresponding policy $\pi_{\boldsymbol\theta}$ maximizes the expected cumulative discounted rewards:
\begin{equation}\label{eq:primal-obj}
J(\boldsymbol\theta) = \mathbb{E}\left[ \sum_{t=0}^T \gamma^t r(s_t, a_t) \right]
\end{equation}
where $T$ is the horizon of an episode.

\subsubsection{Observation space and action space}
As mentioned above, the observation for each robot consists of four parts $[\mathbf{o}_l, \mathbf{o}_v, \mathbf{o}_g, \mathbf{o}_p]$. The 2D laser scans data $\mathbf{o}_l$ consists of 90 measurements from a 360$^\circ$ LiDAR, the velocity $\mathbf{o}_v$ is in Cartesian coordinate, and all of the positions (i.e., $\mathbf{o}_g$ and $\mathbf{o}_p$) are transformed to the local robot coordinate frame from global coordinates.
In our case, the robots are omnidirectional and the control command is the velocity in Cartesian coordinate, i.e., $\mathbf{a} = [v_x, v_y]$. For practical purpose, we shrink the output of the policy $\pi_{\boldsymbol\theta}$ to fit the maximum value of velocity, i.e., $\lVert\mathbf{a}\rVert \leq v_{max}$.   

\subsubsection{Reward design}\label{sec:reward}
The robot team is supposed to get into the target zone without collisions.
To achieve this, the reward function for each robot $i$ is designed to encourage approaching the target zone and avoiding collisions with other robots and obstacles $\mathbf{p}_{obs}$: 
\vspace{-2mm}
\begin{equation}
    r^{i}(s_t,a_t) = r_g^{i,t} + r_c^{i,t}
\end{equation}
where $r_g^{i,t}$ is the reward when the robot approaches the target, or all of the robots enter the target zone:
\begin{align}
r_g^{i, t} =
\begin{cases}
r_{goal}  
&
\begin{aligned}
&\text{if}\; \text{dst}(j,t) < R_g - R_r \\
&\forall j\in [1,N]
\end{aligned}\\
w_g(  \text{dt}(i,t-1)  -  \text{dst}(i,t) ) &\text{otherwise}
\end{cases}
\end{align}
where $\text{dst}(i,t) = \norm{ \mathbf{p}_i^t - \mathbf{g} }$ denotes the distance to goal for robot $i$ at timestep $t$ and $w_g > 0$.
$r_c^{i,t}$ is to penalize collisions among robots or obstacles:
\begin{align}
r_c^{i, t} =
\begin{cases}
r_{coll}  
&
\begin{aligned}
&\text{if} \norm{ \mathbf{p}_i^t - \mathbf{p}_{obs} } < R_r \\
&\text{or} \norm{\mathbf{p}_i^t - \mathbf{p}_j^t} < 2R_r\;\forall j\in[1,N],j\ne i
\end{aligned}\\
0 
&\text{otherwise}
\end{cases}
\end{align}
We empirically set $r_{coll}=-100$, $r_{goal}=100$ and $w_g=10$.
  
\subsection{Global connectivity constraint}\label{sec:conn_constraint}
The approach for global connectivity maintenance of communication networks is based on the second smallest eigenvalue of the associated graph Laplacian, which is briefly described below and used to formulate a connectivity constraint later on.

Consider a multi-agent systems with positions denoted as $\mathbf{p}_i\in\real^2, i\in\{1,2,\dots,N\}$.
Each pair of agents $i$ and $j$ can communicate with each other within an identical limited distance $R_c$,
i.e., $\lVert \mathbf{p}_i - \mathbf{p}_j\rVert\leq R_c$. 
The communication topology can be represented as an undirected
graph $\mathcal{G}=\{ V, E \}$ where each vertex $v_i\in V$ denotes the $i$-indexed agent
and each edge $e_{ij}\in E$ denotes the communication availability between agents $i$ and $j$. 
The Laplacian matrix of graph $\mathcal{G}$ is defined as $L=D-A$ 
where $A\in\real^{N\times N}$ is the adjacency matrix and each element $a_{ij}=1$
if $e_{ij}\in E$ and $a_{ij}=0$ otherwise.
$D\in\real^{N\times N}$ is the degree matrix, 
which is diagonal, i.e., $D= \text{diag}\{ \phi_1, \phi_2,\dots,\phi_N \}$
and the $i$-th element of the diagonal is $\phi_{i}=\sum_{j=1}^N a_{ij}$. 
The second smallest eigenvalue of the Laplacian matrix, denoted as $\lambda_2(\mathcal{G})$, 
describes \textit{algebraic connectivity} of the graph $\mathcal{G}$
as $\lambda_2(\mathcal{G}) \geq 0$ always holds if $\mathcal{G}$ is connected~\cite{fiedler1973algebraic}. 

With the RL formulation, the robot team learns to navigate to the target zone.
To impose global connectivity to the team, we introduce a cost function $c(s_t,a_t):\mathcal{S}\times\mathcal{A}\to\real$ that use the algebraic connectivity measurement::
\begin{equation}
    c(s_t,a_t) = \mathbb{I}(\lambda_2(\mathcal{G}_t) < 0 )
\end{equation}
where $\mathbb{I}$ is the indicator function and $\lambda_2(\mathcal{G}_t)$ is the algebraic connectivity of the graph $\mathcal{G}_t$ which is constructed by $[\mathbf{p}_1^t,\mathbf{p}_2^t,\dots,\mathbf{p}_N^t]$ at timestep $t$. 
Note that the robots can infer $\lambda_2(\mathcal{G}_t)$ from positions of the teammates $\mathbf{o}_p^t$. As global connectivity serves as a constraint, the multi-robot navigation problem is extended to a constrained RL formulation. Specifically, constrained RL aims to optimize the policy while bounding expected discounted returns with respect to the cost function:
\vspace{-2mm}
\begin{equation}\label{eq:second_obj}
J_c(\boldsymbol\theta) = \mathbb{E}\left[ \sum_{t=0}^T \gamma_c^t c(s_t, a_t) \right]
\end{equation}
The Eq.~(\ref{eq:second_obj}) is formed in terms of the expected value of trajectories. Similar to~\cite{pfeiffer2018reinforced}, by setting the discount factor $\gamma_c$ close to 1, we are able to constrain the total number of constraint violations per trajectory to be less than the desired threshold $d$.  
In this work, we use Constrained Policy Optimization (CPO)~\cite{achiam2017constrained} to solve this constrained RL problem:
\vspace{-2.5mm}
\begin{equation}\label{eq:constrained-rl}
\begin{aligned}
     \boldsymbol\theta^\ast =\;& \arg\max  \;J(\boldsymbol\theta) \\
    \text{s.t.} & \;J_c(\boldsymbol\theta) \leq d
\end{aligned}
\end{equation}
where $d$ is the upper bound. CPO adopts trust region methods for local policy search, which is introduced in Trust Region Policy Optimization (TRPO)~\cite{schulman2015trust}. TRPO uses a constraint on the Kullback-Leiber (KL) divergence of two stochastic policies between updates to stabilize policy training with monotonic improvement guarantees. 
Similarly, CPO replaces the original objective and constraint in Eq.~(\ref{eq:constrained-rl}) with surrogate functions in a trust region:
\vspace{-2.5mm}
\begin{equation}\label{eq:cpo-problem}
\begin{aligned}
 \boldsymbol\theta^{k+1} =\;& \arg\max_{\boldsymbol\theta} \;\mathbb{E}_{s,a\sim d^{\pi_{k}}}\left[\frac{\pi_{\boldsymbol\theta}(s)}{\pi_k(s)} A^{\pi_k}(s,a) \right] \\
    \text{s.t.} & \; J_c(\boldsymbol\theta^k) + \mathbb{E}_{s,a\sim d^{\pi_k}}\left[ \frac{\pi_{\boldsymbol\theta}(s)}{\pi_k(s)}  \frac{A_c^{\pi_k}(s,a)}{1-\gamma_c} \right] \leq d   \\
    & \;\bar{D}_{KL}(\pi_{\boldsymbol\theta} || \pi_{k}) \leq \eta
\end{aligned}
\end{equation}
\vspace{-0.5mm}
where $\bar{D}_{KL}(\pi_{\boldsymbol\theta} || \pi_{\boldsymbol\theta_k})$ is the average KL divergence between two consecutive policies and $\eta > 0$ is the step size.
Advantages functions $A^{\pi_k}(s,a),A_c^{\pi_k}(s,a)$ are introduced in Eq.~(\ref{eq:cpo-problem}) which can be estimated by function approximators $V_{\boldsymbol\phi}, V_{\boldsymbol\phi_c}$ with one-step TD error
$A(s,a) = r(s,a) + \gamma V(s_{t+1}) - V(s_t)$.
CPO approximately solves the surrogate optimization in Eq.~(\ref{eq:cpo-problem}). Due to approximation errors, CPO asymptotically bounds the constraint during training.
One may refer to \cite{achiam2017constrained} for details.

\subsection{Behavior cloning}\label{sec:bc}
In fact, the optimization problem in Eq.~(\ref{eq:constrained-rl}) regarding two objectives (i.e., Eq.~(\ref{eq:primal-obj}) and Eq.~(\ref{eq:second_obj})) is usually infeasible as it leads to conservative policy explorations in stochastic domains even without a trust region constraint. In our case, connectivity constraint leads to in-place oscillations, and the robots can not learn to navigate to the target. 

To make this policy optimization tractable, we introduce the behavior cloning (BC) technique during RL training. 
In many robotic tasks such as manipulation, limited demonstration data can be collected, but an expert policy for all states is usually infeasible~\cite{vecerik2017leveraging,rajeswaran2017learning}. 
However, for map-less navigation, recent successful practices have shown that an expert can be obtained by RL methods without great effort. 
In particular, given observation $\tilde{\mathbf{o}} = [\mathbf{o}_l, \mathbf{o}_v, \mathbf{o}_g]$ and the same reward structure ($N=1$) in Sec\ref{sec:reward}, RL (we use Soft Actor-Critic~\cite{haarnoja2018soft} here) can train a navigation expert policy $\pi_E(\tilde{\mathbf{o}})$ for a single robot efficiently.
Though BC is widely used for pre-training to provide initialization for RL. As a result, pre-training from a navigation expert can not provide a good bootstrap for this problem since it performs undesired behaviors that break the global connectivity. 
Instead, we incorporate BC into RL training.
With a navigation expert $\pi_E$, a supervised mean square error (MSE) loss function is constructed to minimize the difference with $\pi_{\boldsymbol\theta}$:
\vspace{-1.5mm}
\begin{equation}\label{eq:bc}
    J_{BC}(\boldsymbol\theta) = 
    \frac{1}{T}\sum_{t=0}^T 
\norm{    \pi_E(\tilde{\mathbf{o}}^t) - \pi_{\boldsymbol\theta}(\mathbf{o}^t)  }^2
\end{equation}
Then the original RL objective Eq.~(\ref{eq:primal-obj}) is modified with this additional BC loss term:
\vspace{-1.5mm}
\begin{equation}\label{eq:new_obj}
    \tilde{J}(\boldsymbol\theta) = J(\boldsymbol\theta) - \lambda_e J_{BC}(\boldsymbol\theta)
\end{equation}
where $\lambda_e > 0$ is a hyper-parameter.

As a result, the robots attempt to mimic the expert behaviors when optimizing the objective Eq.~(\ref{eq:new_obj}) but eventually surpass the performance of expert $\pi_E$ due to improvements from the policy gradients of the RL term Eq.~(\ref{eq:primal-obj}).
\subsection{Neural network structure}
\begin{figure}[t]
    \centering
    \begin{subfigure}{0.25\textwidth}
    \centering
    \includegraphics[width=0.85\textwidth]{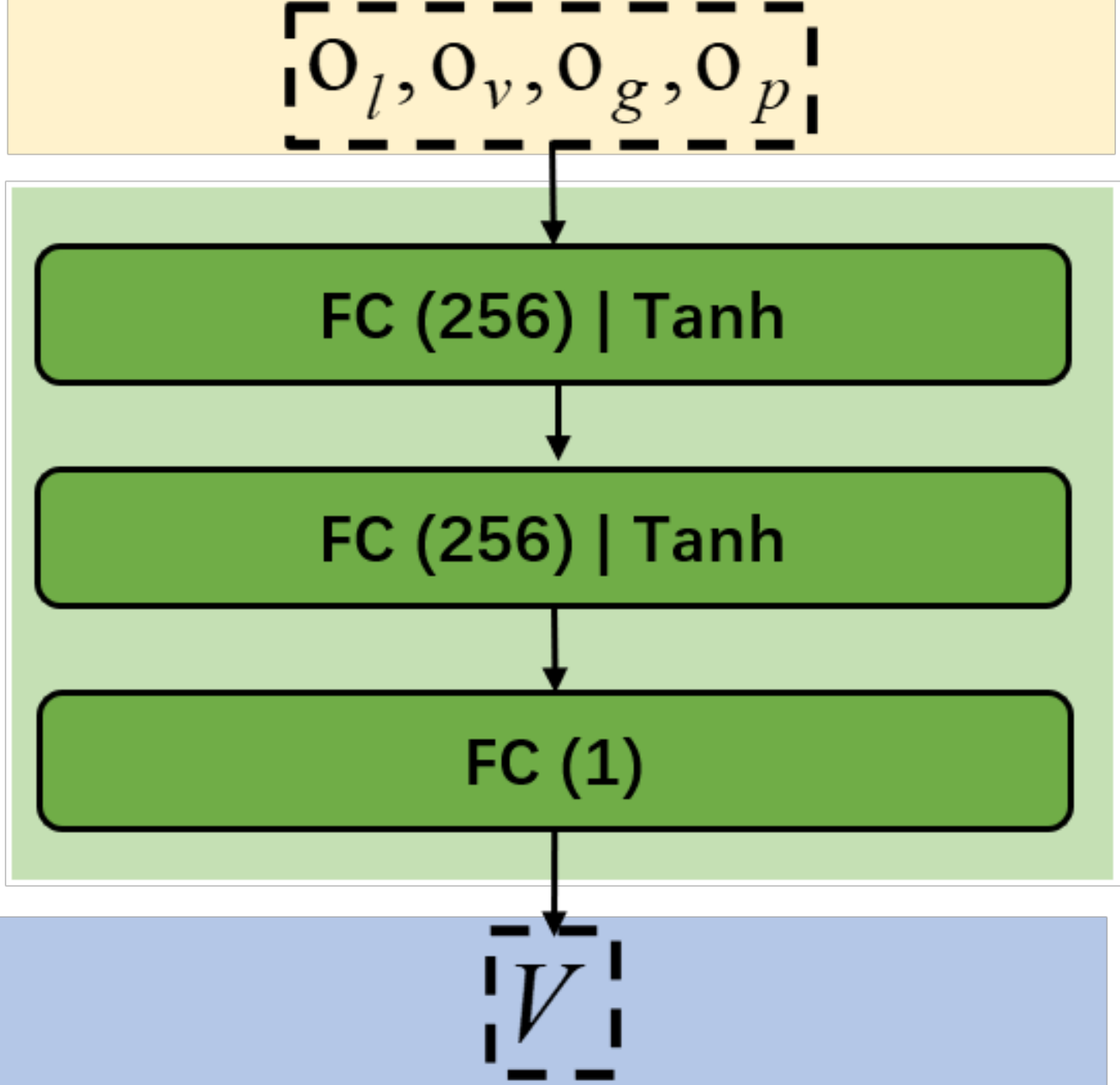}
    \caption{Value network}
    \end{subfigure}%
    \begin{subfigure}{0.25\textwidth}
    \centering
    \includegraphics[width=0.95\textwidth]{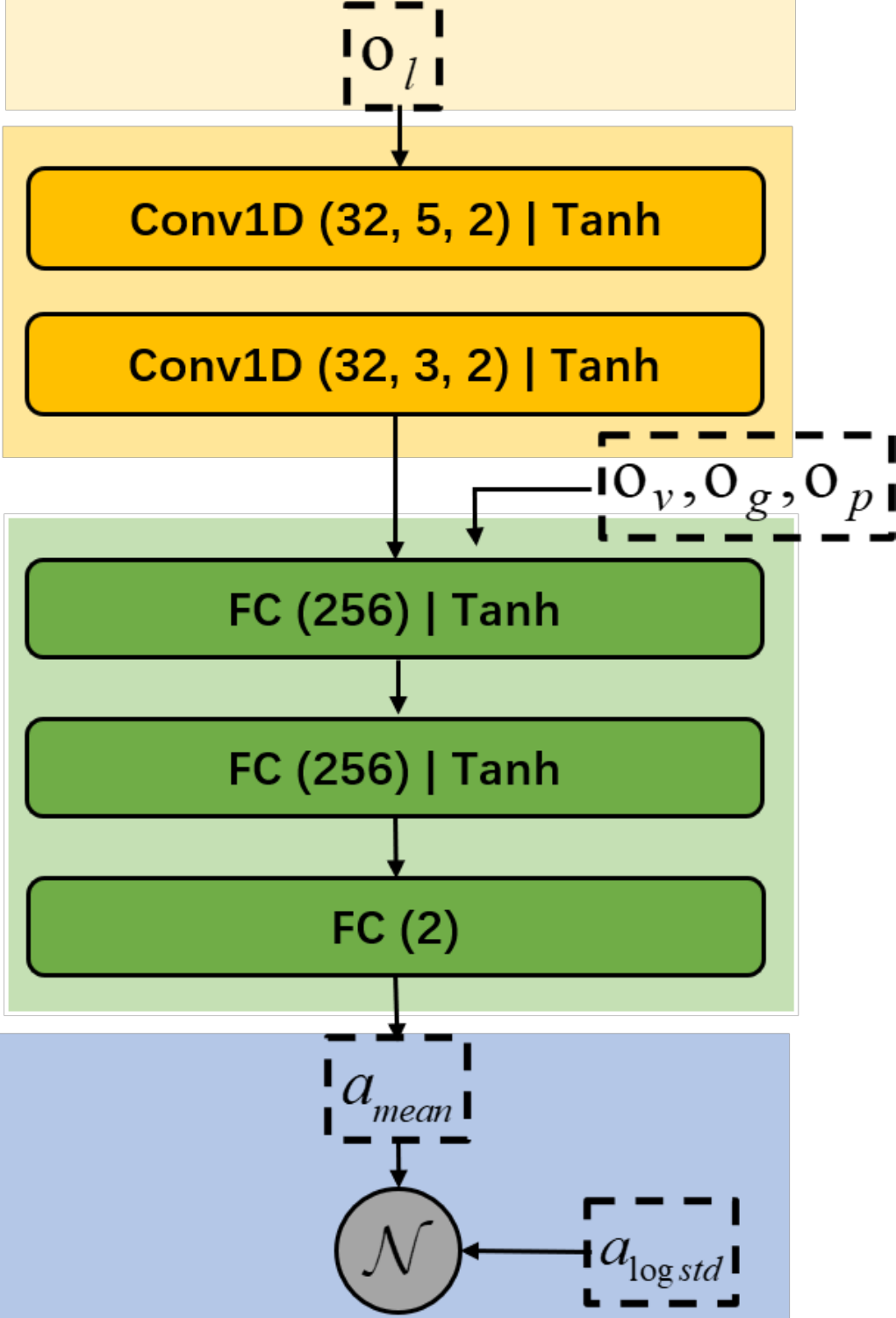}
    \caption{Policy network}%
    \end{subfigure}
    \caption{Network structure. Conv1D (filters, kernel, stride) denotes the 1D convolutional layer, FC denotes the fully-connected layer.}
    \vspace{-5mm}
    \label{fig:network}
\end{figure}

We follow the common practices~\cite{long2018towards,lin2019end, zhang2020map} in the parameterization of policy $\pi_{\boldsymbol\theta}$ and adopt a similar network structure as shown in Fig.~\ref{fig:network}. The raw laser data $\mathbf{o}_l$ is processed by two 1D convolution (\texttt{Conv1D}) layers and a fully connected (\texttt{FC}) layer for feature extraction. The output feature vector is concatenated with the rest of observation vector $[\mathbf{o}_v, \mathbf{o}_g, \mathbf{o}_p]$ and are fed into two hidden \texttt{FC} layers. Then the velocity output $\mathbf{a}_{mean}$ is obtained from the last \texttt{FC} layer. As suggested by a large scale experimental results~\cite{engstrom2019implementation}, we use hyperbolic tangent (\texttt{tanh}) as activation functions and orthogonal initialization scheme to initialize network weights.    
The output $\mathbf{a}_{mean}$ is supposed to be the mean of the action distribution. A separate vector $\mathbf{a}_{logstd}$ with the same dimension as $\mathbf{a}_{mean}$ is defined as the log standard deviation that will be updated alone. In this way, the stochastic policy $\pi_{\boldsymbol\theta}$ is formed as a Gaussian distribution $\mathcal{N}\left(mean=\mathbf{a}_{mean}, std=\exp(\mathbf{a}_{logstd})\right)$. At each timestep $t$, the robots take random actions $\mathbf{a}^t\sim \pi_{\boldsymbol\theta}(\mathbf{o}^t)$ for on-policy exploration.

The value networks $V_{\boldsymbol\phi}$ and $V_{\boldsymbol\phi_c}$ are introduced for advantage estimation and both constructed with two hidden and one output \texttt{FC} layers to map the observation $\mathbf{o}$ to value estimates.

\subsection{Policy training}
\subsubsection{Simulation setup}
\begin{figure}[h]
    \begin{subfigure}{0.12\textwidth}
    \centering 
    \includegraphics[width=\textwidth]{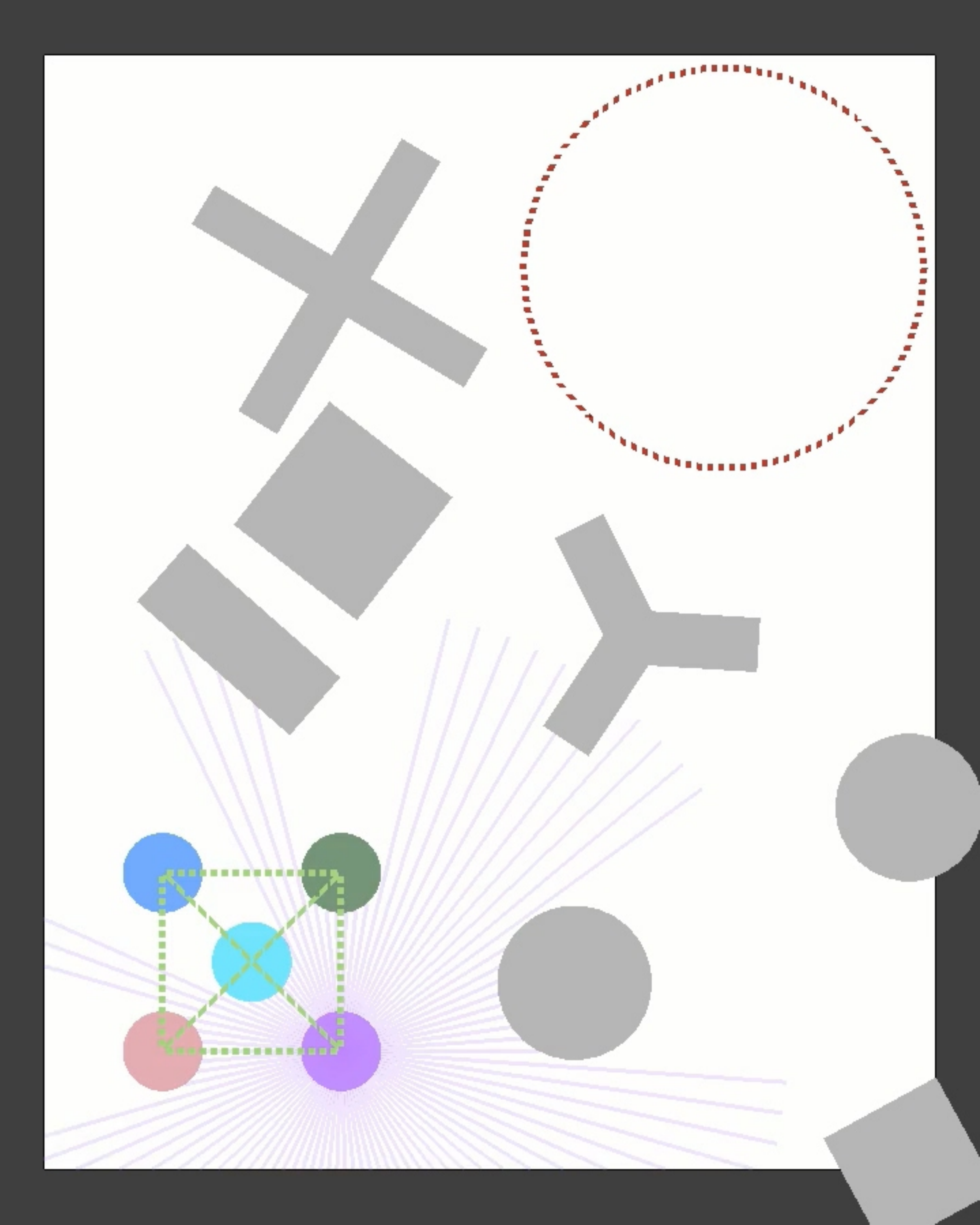}
    \end{subfigure}\hspace{\fill}%
    \begin{subfigure}{0.12\textwidth}
    \centering
    \includegraphics[width=\textwidth]{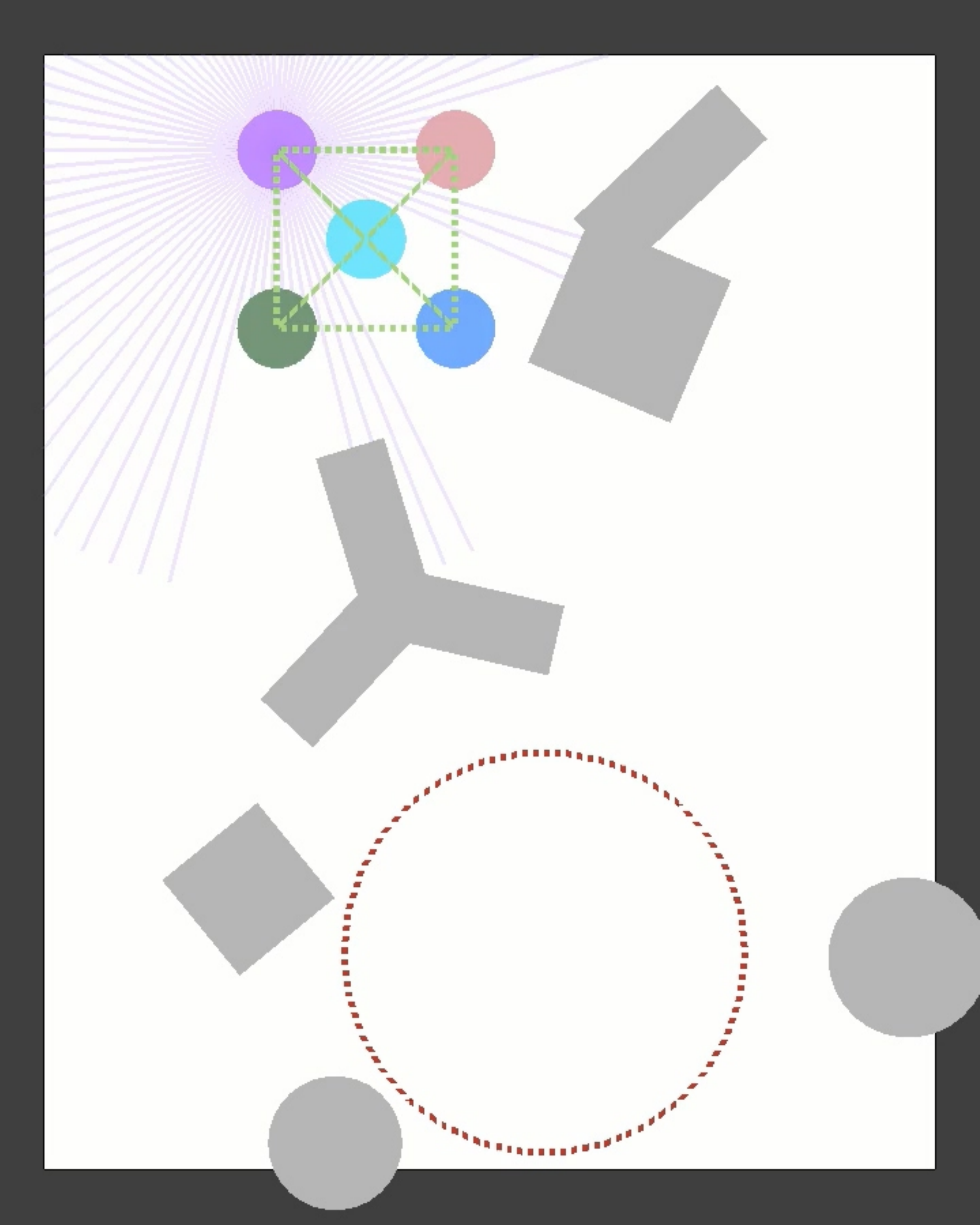}
    \end{subfigure}\hspace{\fill}%
    \begin{subfigure}{0.12\textwidth}
    \centering
    \includegraphics[width=\textwidth]{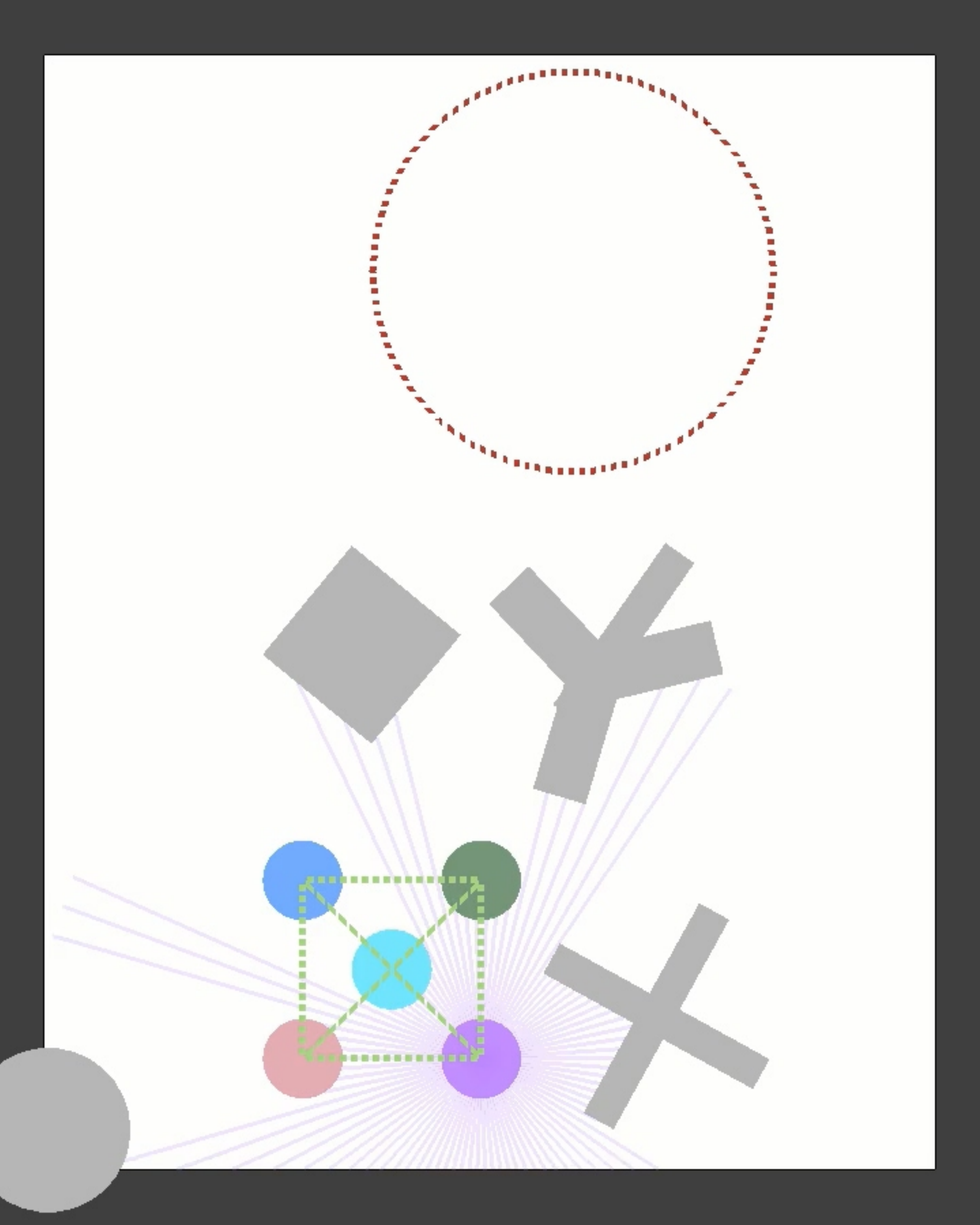}
    \end{subfigure}\hspace{\fill}%
    \begin{subfigure}{0.12\textwidth}
    \centering
    \includegraphics[width=\textwidth]{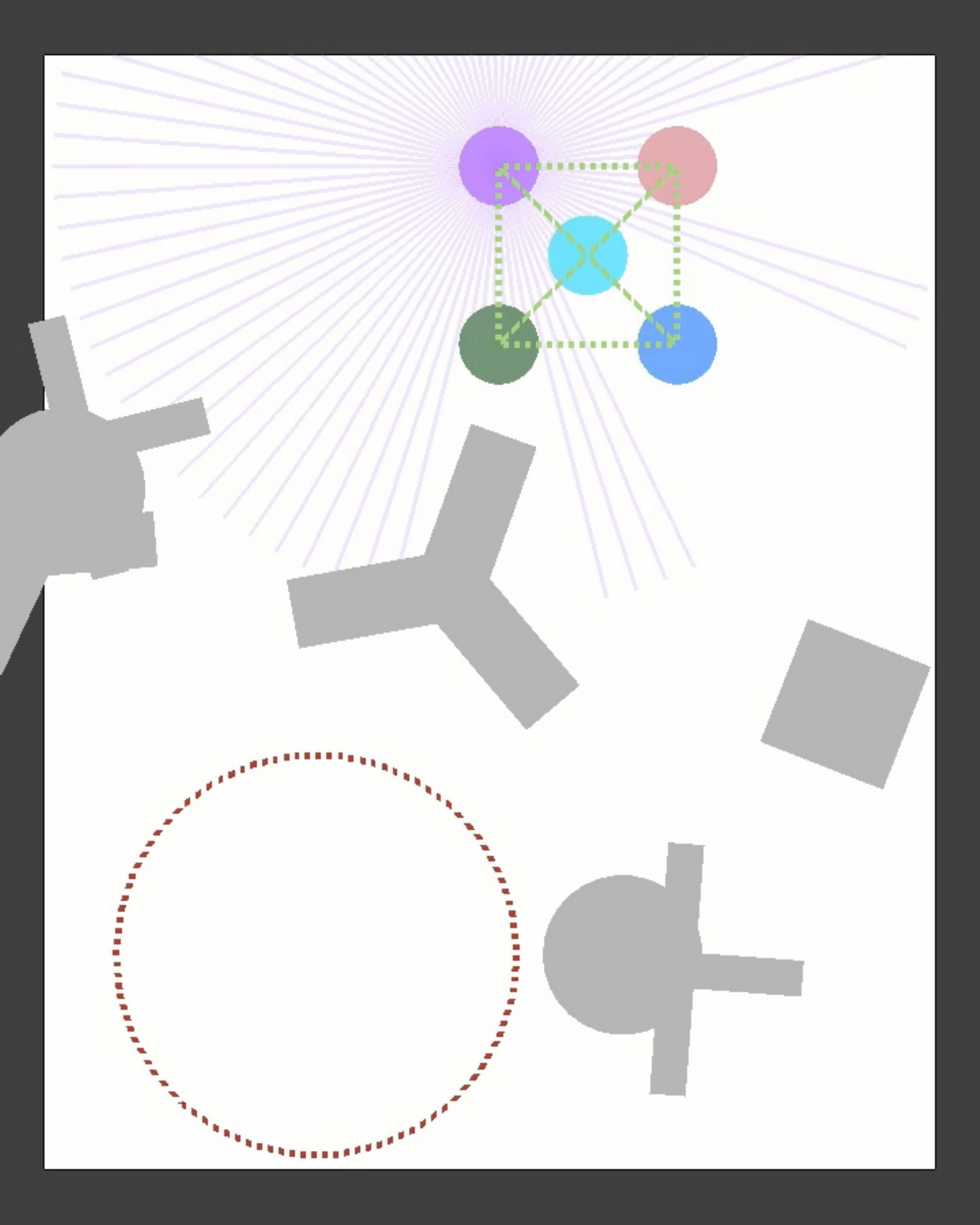}
    \end{subfigure}
    \caption{ Selected simulation scenarios. Simulation configurations are randomly generated. The robots (colorful circles) aim to get into the target (red dashed line) and avoid obstacles (gray). The green lines represent connectivity links between robots. }
    \label{fig:simulation-scenes}
\end{figure}

Since RL is considerably sample inefficient, the expert policy $\pi_E$ and the shared policy $\pi_{\boldsymbol\theta}$ are trained in the simulation where sufficient samples are available. The multi-robot navigation is simulated in a custom OpenAI Gym environment~\cite{1606.01540}. We use the domain randomization technique~\cite{tobin2017domain} in order to avoid overfitting and bridge the sim-to-real gap. Concretely, we build 100 2D maps in total where several elements are randomly initialized, including target positions, starting points of robots, and obstacle configurations (numbers, shapes, positions, and sizes) as shown in Fig.~\ref{fig:simulation-scenes}. 
\vspace{-2mm}

\hypersetup{linkcolor=black}

\hypersetup{linkcolor=RoyalPurple}
\vspace{-3mm}
\subsubsection{CPO policy update}
The policy training algorithm is shown in Algorithm~\ref{algo:cpo}.
The robots collect transition data simultaneously in several maps that are randomly selected. At each timestep $t$, each robot takes actions generated from the shared policy $\pi_{\boldsymbol\theta}$ given its own observation $\mathbf{o}^{i,t}$ and then receives its immediate reward $r^{i,t}$ and a shared connectivity cost signal $c^t$. 
When the number of collected transition data is sufficient (i.e., greater than $N_{batch}$), we use the batch of episodes collected by all robots to update policy $\pi_{\boldsymbol\theta}$. 
To build up CPO formulation,
the objective and the constraint are approximated with advantage estimates (line \ref{line:adv}). We then use the well-behaved expert $\pi_E$, which is trained in advance to compute the supervised BC loss with all transition data (line \ref{line:bc}). The policy network $\pi_{\boldsymbol\theta}$ is optimized by CPO with regard to the combined objective in Eq. (\ref{eq:new_obj}). 
The value networks $V_{\boldsymbol\phi}$ is optimized with the following loss function:
\begin{equation}\label{eq:value_loss}
L_V(\boldsymbol\phi) = 
\sum_{t=1}^T \norm {V_{\boldsymbol\phi}(s_t) -  \sum_{l=0}^\infty \gamma^t r(s_{t+l},a_{t+l})  }^2
\end{equation}
Eq.~(\ref{eq:value_loss}) is computed with all sampled trajectories and minimized with an L-BFGS optimizer.
The cost value network $V_{\boldsymbol\phi_c}$ is updated in the same fashion in terms of cost signals.
\begin{algorithm}[t]
\begin{algorithmic}[1]
\State Apply orthogonal initialization to policy network $\pi_{\boldsymbol\theta}$ and value networks $V_{\boldsymbol\phi}, V_{\boldsymbol\phi_c}$, provide expert policy $\pi_E$
\State $\psi_{s} = 1$
\For{ episode $\psi=1,2,\dots$ }
\State Receive initialized observations $\{ \mathbf{o}^{i,t} \}_{i=1}^N$ from the environment
\For{robot $i=1,2,\dots,N$ }
\State Take action $\mathbf{a}^{i,t}\sim \pi_{\boldsymbol\theta}(\mathbf{o}^{i,t})$\label{line:sample}
\State Receive reward $r^{i,t}$ and cost signal $c^t$ \label{line:r_c}
\State Repeat \ref{line:sample} and \ref{line:r_c}, collect transition data $(\mathbf{o}^{i,t},\mathbf{a}^{i,t},r^{i,t},c^t)$ for $T_\psi$ timesteps
\EndFor
\hypersetup{linkcolor=RoyalPurple}
\If{ $\sum_{j=\psi_{s}}^\psi T_j > N_{batch}$ }
\State // \textit{Update $\pi_{\boldsymbol\theta}$ and $V_{\boldsymbol\phi}, V_{\boldsymbol\phi_c}$ in batch } 
\State Compute advantage estimates for batch data
$r^{i,t} + \gamma V_{\boldsymbol\phi}(\mathbf{o}^{i+1,t}) - V_{\boldsymbol\phi}(\mathbf{o}^{i,t})$,
$c^{t} + \gamma V_{\boldsymbol\phi_c}(\mathbf{o}^{i+1,t}) - V_{\boldsymbol\phi_c}(\mathbf{o}^{i,t})$\label{line:adv}
\State Compute BC loss term Eq. (\ref{eq:bc}) with expert policy $\pi_E$ and add to the surrogate objective\label{line:bc}
\State Update $\boldsymbol\theta$ by approximately solving CPO in Eq. (\ref{eq:cpo-problem})
\State Update $\boldsymbol\phi, \boldsymbol\phi_c$ by minimizing Eq. (\ref{eq:value_loss}) with L-BFGS~\cite{liu1989limited}
\EndIf{}
\State $\psi_s = \psi$
\hypersetup{linkcolor=black}
\EndFor
\end{algorithmic}
\caption{Policy Training with CPO and BC}
\label{algo:cpo}
\end{algorithm}

We implement our neural network models in PyTorch. All training procedures and simulation experiments are performed on a computer with an Nvidia RTX 2080 GPU and an Intel Xeon Gold 6254 CPU. 
Each timestep takes about 2 $\mathrm{ms}$ in various simulations. 
The primal hyper-parameters for Algorithm~\ref{algo:cpo} are shown in Tab.~\ref{tab:param}.
\vspace{-3mm}
\begin{table}[h]    
\caption{Key parameter settings for training}
    \label{tab:param}
    \centering
    \begin{tabular}{c|c|c}
    \hline
    \hline
        Parameter & Description & Value \\
        \hline
        $N_{batch}$ & number of timesteps for each update & 2048 \\
        $\gamma$ & discount factor for reward & 0.99 \\
        $\gamma_c$ & discount factor for cost & 0.999 \\
        $\lambda_e$ & coefficient for BC loss & 0.1 \\
        $lr_\phi$, $lr_{\phi_c}$ & L-BFGS step size for $V_{\boldsymbol\phi}, V_{\boldsymbol\phi_c}$ & 0.1 \\
        $\eta$ & KL constraint threshold & 0.01 \\
        $d$ & connectivity constraint threshold & 0.1 \\
        \hline
        \hline
    \end{tabular}

\end{table}
\vspace{-4mm}
\section{Experiments and results}
To demonstrate the effectiveness of our approach, we first investigate the influences of behavior cloning and the connectivity constraint on policy performance. 
Then we evaluate the performance of our approach in real-world experiments.\begin{table}[t]
\caption{Evaluation of generalization performance on various approaches with the different number of robots. The results are evaluated in 100 unseen scenarios.}
    \label{tab:performance}
    \centering
    \begin{tabular}{c|c|c|c|c}
    \hline
    \hline
      Metrics & Approach & 3 robots & 4 robots & 5 robots \\
     \hline
\multirow{3}{*}{Success Rate} & CPO     & 0.05 & 0.02 & 0.03\\                              & TRPO+BC & 0.9  & 0.95 & 0.95 \\
                              & CPO+BC  & 0.87 & 0.88 & 0.82\\
      \hline
\multirow{3}{*}{Connectivity Rate} & CPO     & 0.97 & 0.98 & 0.97\\                                   
& TRPO+BC & 0.3  & 0.13 & 0.07\\
                                   & CPO+BC  & 0.73 & 0.77 & 0.71\\
      \hline
\multirow{3}{*}{Travel Time} & CPO     & 16.68 & 14.732 & 10.307\\                             & TRPO+BC & 7.853 & 9.564 & 11.421\\
                             & CPO+BC  & 7.28 & 8.697 & 9.163\\
      \hline
      \hline
    \end{tabular}
\vspace{-3mm}
\end{table}
\vspace{-4mm}
\subsection{Comparisons on various approaches in simulation}
To study the influences of behavior cloning and connectivity constraint in our approach, we compare three approaches with different combinations:
\begin{itemize}
    \item \textit{CPO}: policy optimization under the connectivity constraint without supervised BC loss term, i.e., the BC loss coefficient $\lambda_e=0$
    \item \textit{TRPO+BC}: unconstrained policy optimization with supervised BC loss term where the connectivity constraint is removed.
    \item \textit{CPO+BC (our approach)}: policy optimization under the connectivity constraint and the supervised BC loss term
\end{itemize}
The three approaches are compared in terms of average rewards and constraint value during training.
To quantitatively measure the final performance of the approaches, we use the following metrics:
\begin{itemize}
    \item \textit{Success Rate}: the ratio of the robot team to reach targets without collisions
    \item \textit{Connectivity Rate}: the ratio of the robot team to maintain connectivity during navigation
    \item \textit{Travel Time}: the average travel time for the robot team successfully navigates to targets
\end{itemize}

\begin{figure*}
\resizebox{\textwidth}{!}{
   \begin{subfigure}{0.23\textwidth}
    \centering 
    \includegraphics[width=\textwidth]{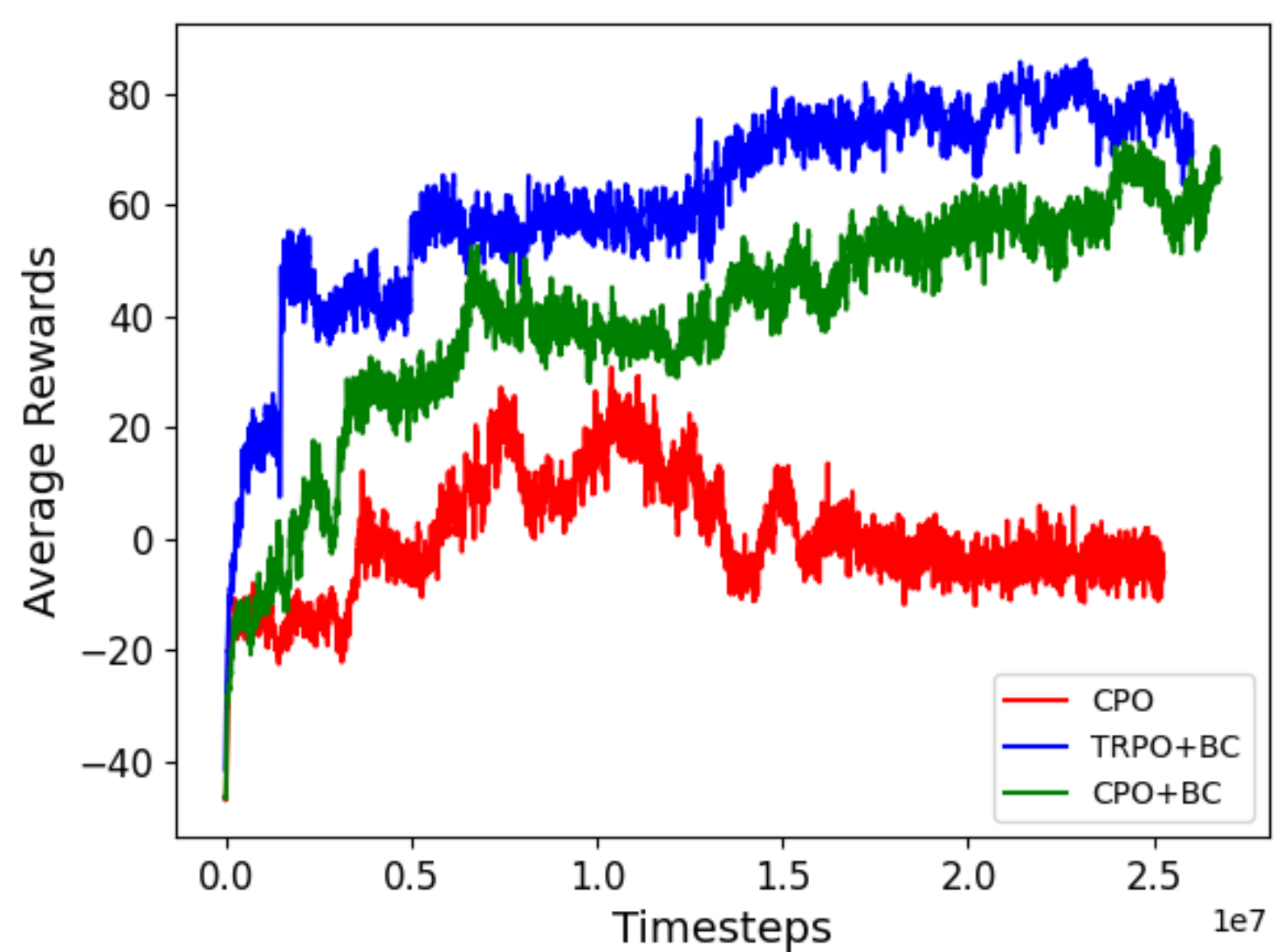}
    \caption{}%
    \label{fig:avg-reward}
    \end{subfigure}
    \begin{subfigure}{0.23\textwidth}
    \centering 
    \includegraphics[width=\textwidth]{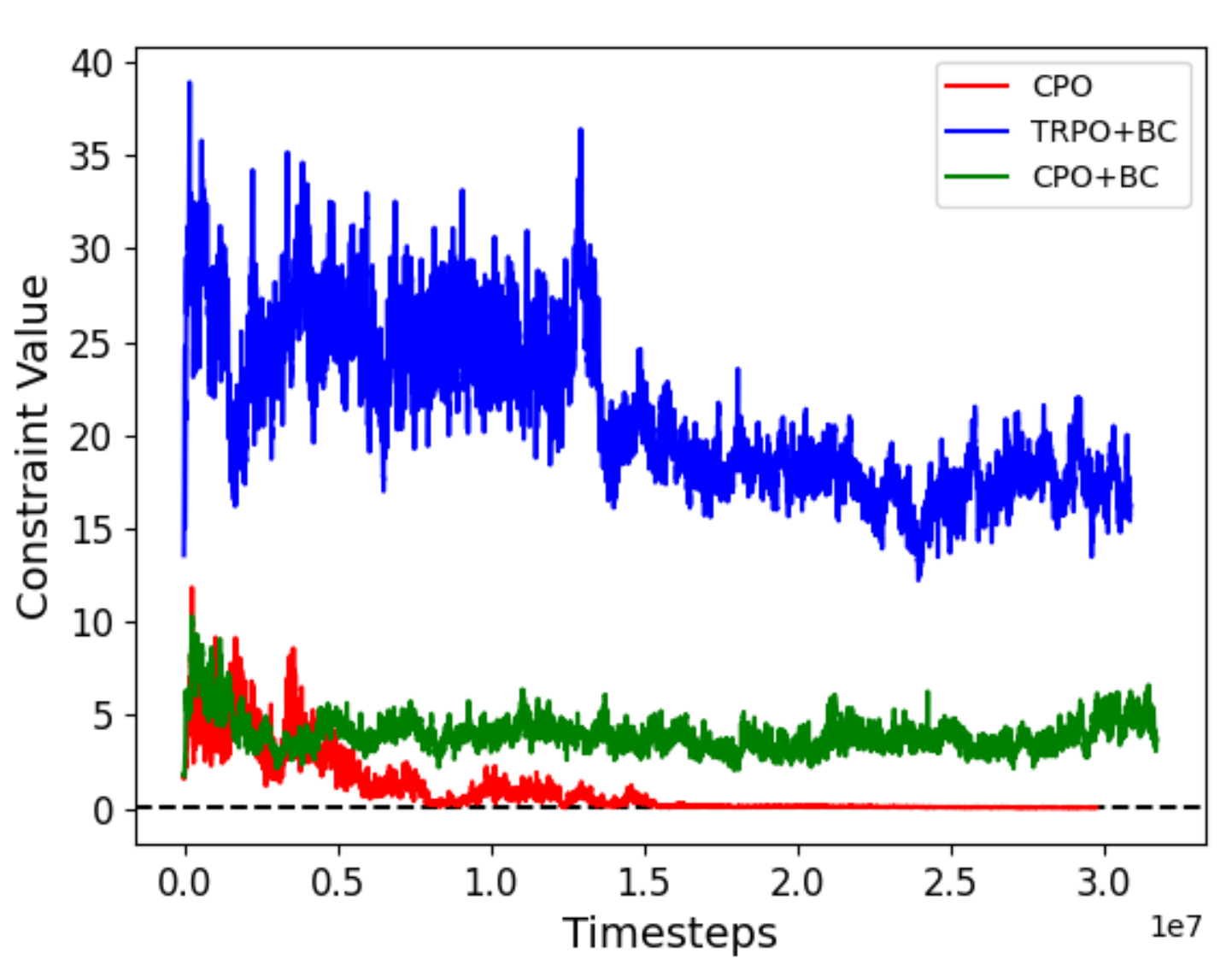}
    \caption{}%
    \label{fig:constraint-val}
    \end{subfigure}
    \begin{subfigure}{0.23\textwidth}
    \centering 
    \includegraphics[width=\textwidth]{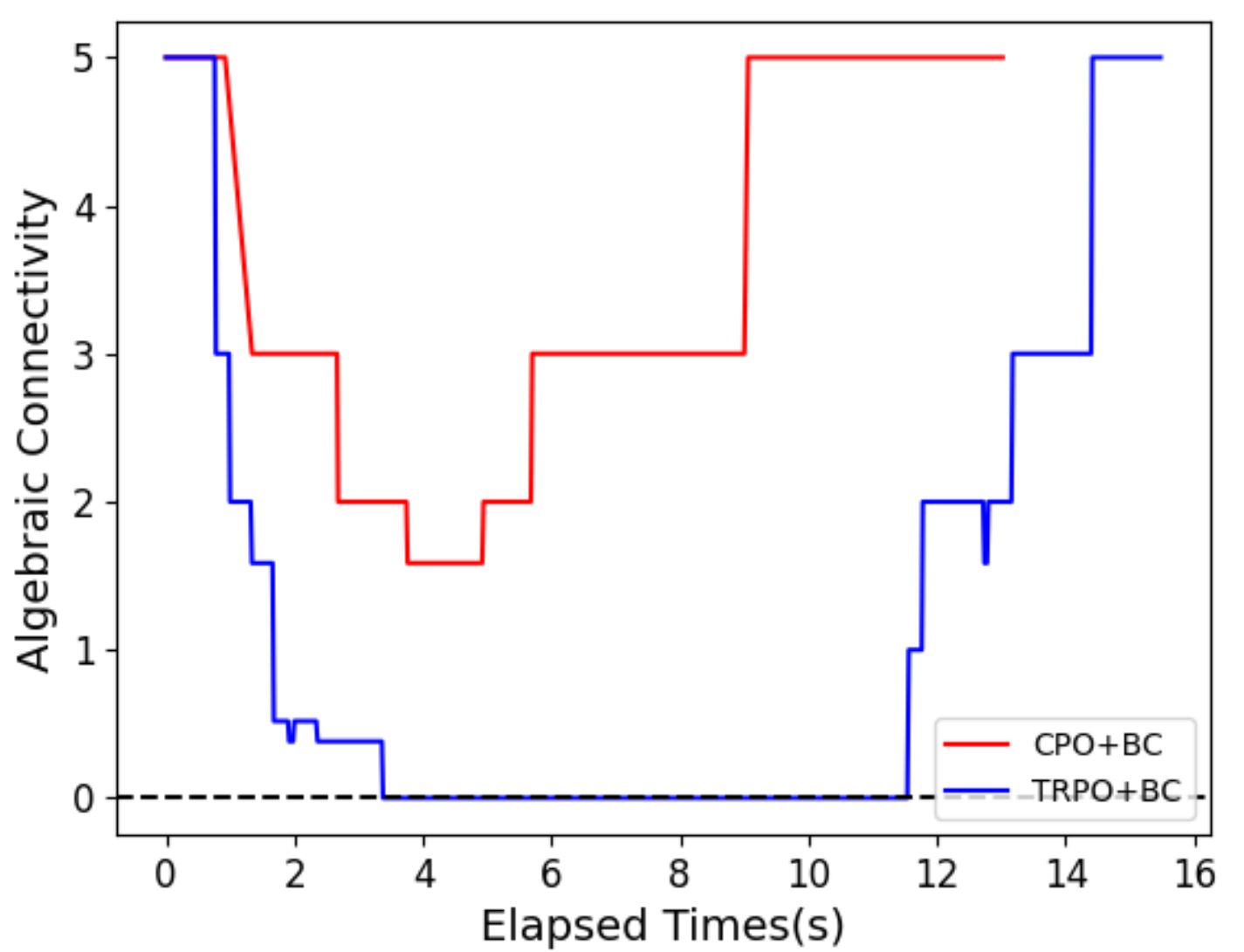}
    \caption{}%
    \label{fig:conn-real-world}
    \end{subfigure}
}
\caption{(a) The evolution of average rewards throughout the training process of various approaches. (b) The evolution of constraint value throughout the training process of various approaches. The black dashed line indicates the predefined threshold of constraint values. (c) Algebraic connectivity of the robot team in real-world experiments. The dashed line represents the connectivity threshold.}
\vspace{-7mm}
\label{fig:my_label}
\end{figure*}

 We randomly re-generate 100 maps for final performance evaluation. We conduct experiments with the various number of robots in simulation and evaluate the performance on average with 5 random seeds for each approach. The evolution of learning curves is consistent across different simulation settings, and we present one of the results (3 robots) in Fig.~\ref{fig:avg-reward} and Fig.~\ref{fig:constraint-val}. 

\textit{Influence of behavior cloning}:
Comparing $\textit{CPO}$ and $\textit{CPO+BC}$ shows that behavior cloning  help to improve the policy convergence speed, as shown in Fig.~\ref{fig:avg-reward}. In fact, \textit{CPO} fails to learn to navigate in most of the training maps because of influences of the connectivity constraint. This is supported by the constraint value evolution over time in Fig.~\ref{fig:constraint-val}. As shown in Tab.~\ref{tab:performance}, the \textit{CPO} policy leads to terrible failures in terms of navigation success rate and dedicates to preserve connectivity in testing scenarios, which are undesired behaviors for navigation tasks. 
\textit{CPO+BC} stabilizes the constraint value to be close to the threshold and succeeds in the majority of unseen scenarios. We conclude that BC has significant effects on policy improvement.

\textit{Influence of connectivity constraint}:
Comparing \textit{CPO+BC} and \textit{TRPO+BC} shows that \textit{CPO+BC} is inferior in terms of policy convergence speed. This is due to the fact that \textit{CPO+BC} learns to satisfy the constraint. However, \textit{TRPO+BC} leads to
high constraint values without bounding the constraint on training convergence. The final performance presented in Tab.~\ref{tab:performance} also proves that \textit{TRPO+BC} fails to maintain connectivity in most cases even it has good navigation performance. Surprisingly, \textit{CPO+BC} outperforms \textit{TRPO+BC} at travel time. We believe that it is because \textit{CPO+BC} learns implicit cooperative schemes as presented in many cases.

 Compared to \textit{TRPO+BC}, \textit{CPO+BC} leads to sub-optimal navigation success rate but we find that most of the failures are caused where the team struggles in complex scenarios which require great exploration abilities. This result indicates that enforcing connectivity constraints degrades the ability of navigation exploration. Compared with CPO, TRPO allows to take "risky" actions without  concerns of constraints satisfaction. Both \textit{CPO+BC} and \textit{TRPO+BC} show consistent performance across the various number of robots in the experiment results listed in Tab.~\ref{tab:performance}.
\vspace{-10mm}
\subsection{Real-world experiments}
\vspace{-4mm}
As \textit{CPO+BC} and \textit{TRPO+BC} have good navigation performance in simulation, we deploy the policies trained by both methods to evaluate their sim-to-real capabilities. The deployed policies take around 2M training timesteps.

\textit{Real robot configuration}:
Our holonomic robot model is shown in Fig.~\ref{fig:robot}. 
The robot is equipped with DJI Mecanum wheel chassis for omnidirectional mobility and an RPLIDAR A2 laser range scanner. Each LiDAR on a robot is deployed at a different height to detect each other. We evenly pick 90 measurements from every 360$^\circ$ laser scan.
The robot is controlled by a Jetson TX2 module that publishes the policy output at 10Hz.
For both the simulations and the real-world experiments we set 
the radius of the robot $R_r=0.18\mathrm{m}$ and the maximum moving speed $v_{max}=0.7\mathrm{m}/\mathrm{s}$. We set $R_c=1.2\mathrm{m}$ to emulate limited communication range.
To distribute position information to the robots, we use an OptiTrack global tracking system to capture and inform the states of robots and the target position.
The 
robots receive laser data and position information by subscribing to ROS nodes of LiDAR and OptiTrack.
\begin{minipage}[!t]{0.48\textwidth}
\begin{minipage}[h]{0.5\textwidth}
\begin{figure}[H]   
    \includegraphics[width=\textwidth]{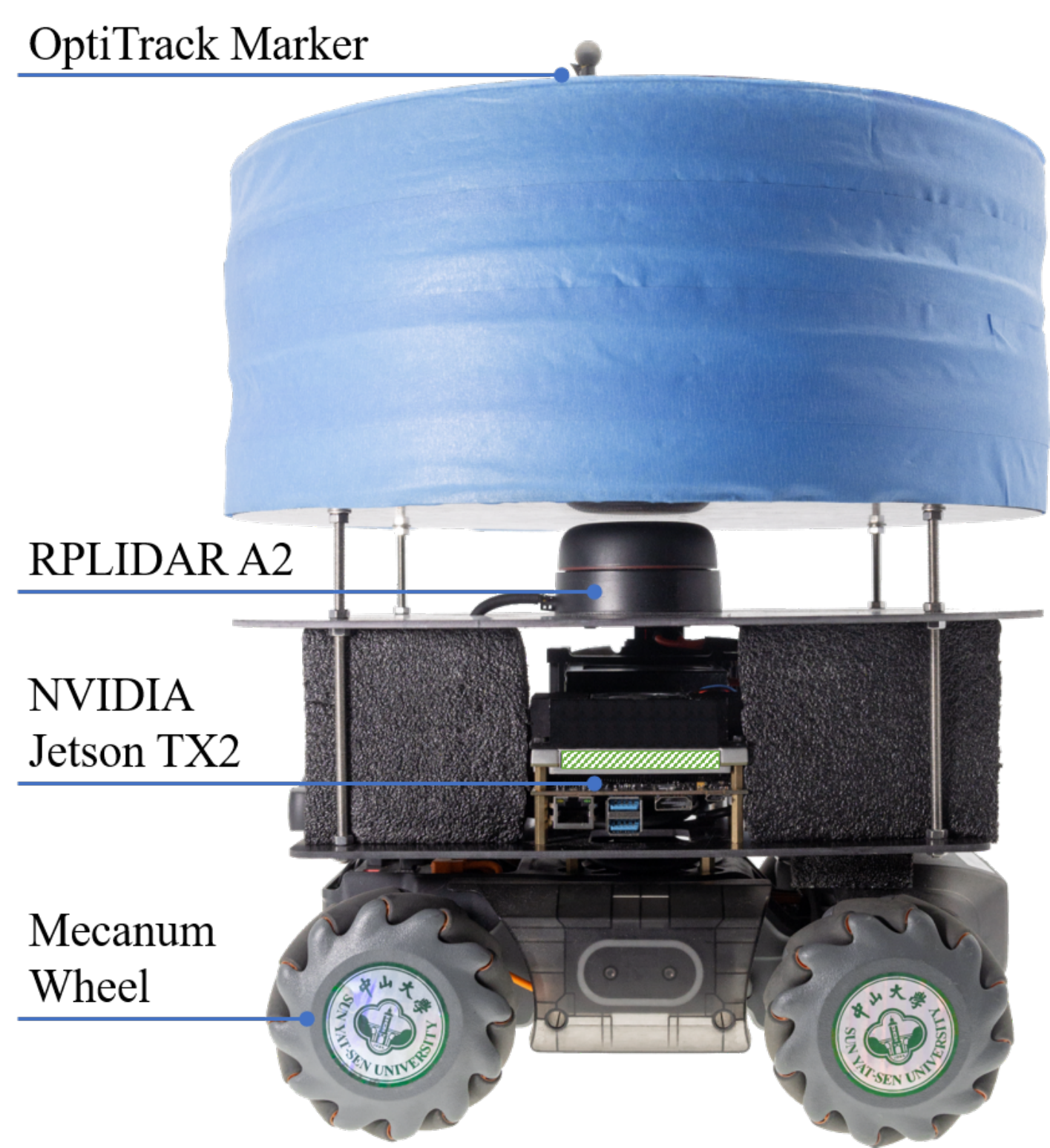}
    \caption{Illustration of the robot.}
    \label{fig:robot}
\end{figure}    
\end{minipage}\hspace{\fill}
\begin{minipage}[h]{0.44\textwidth}
\begin{figure}[H]
    \begin{subfigure}{0.92\textwidth}
    \centering 
    \includegraphics[width=\textwidth]{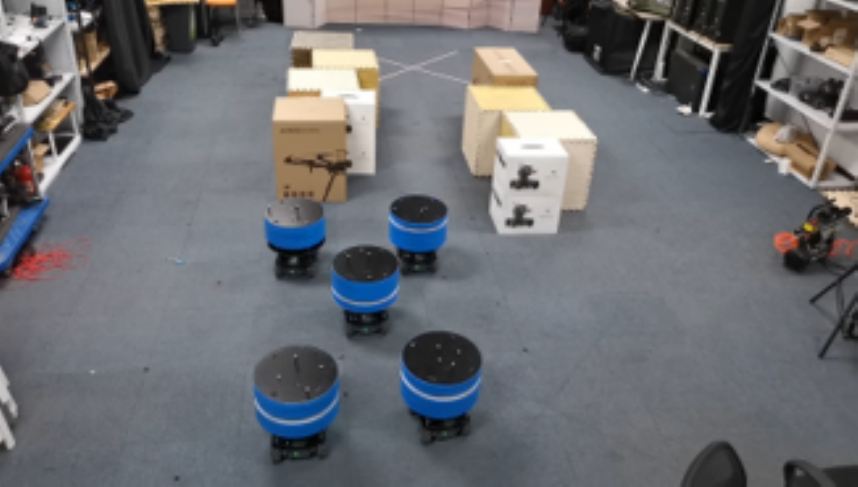}
    \end{subfigure}\vspace{1mm}
    \begin{subfigure}{0.92\textwidth}
    \centering
    \includegraphics[width=\textwidth]{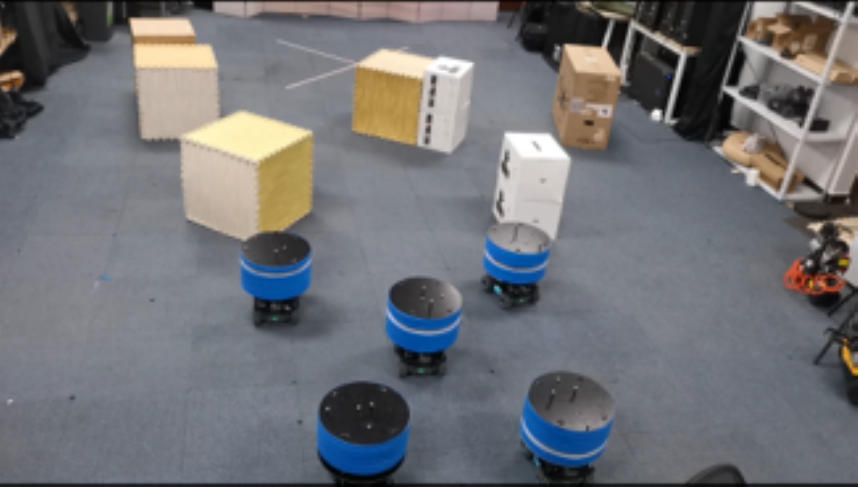}
    \end{subfigure}\vspace{1mm}
    \begin{subfigure}{0.92\textwidth}
    \centering
    \includegraphics[width=\textwidth]{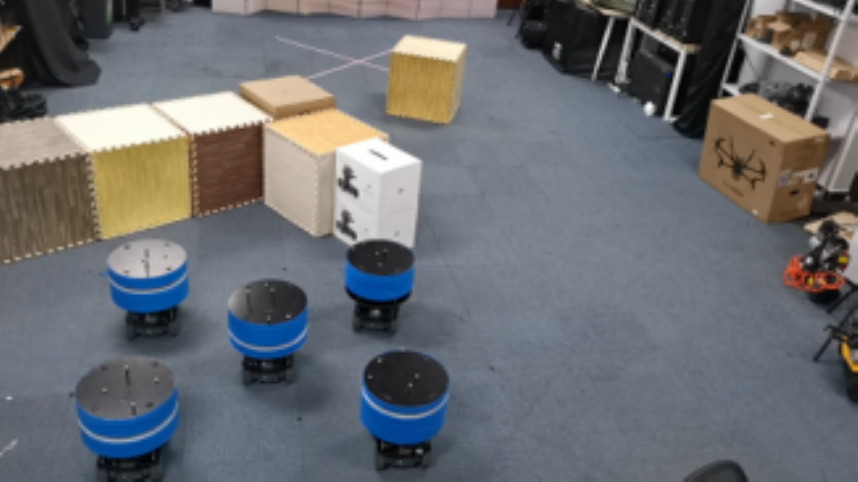}
    \end{subfigure}
    \caption{ Real-world scenarios. }
    \label{fig:scene}
\end{figure}
\end{minipage}
\vspace{3mm}
\end{minipage}

The policies are tested in various scenarios (shown in Fig.~\ref{fig:scene}) and show consistent performance as in simulation on the three metrics in Tab.~\ref{tab:performance}. The experiment results are summarized in Fig.~\ref{fig:conn-real-world}, which shows the value of algebraic connectivity of the robots, as the multi-robot system evolves. Our approach (i.e., \textit{CPO+BC}) takes less travel time and maintains connectivity at all time while \textit{TRPO+BC} fails in connectivity perseverance (the value goes to 0). 

\vspace{-1.5mm}
\section{Conclusion}
In this work, we present an RL-based approach to develop a decentralized policy for multi-robot navigation with connectivity maintenance. 
The policy is shared among robots and jointly optimized by transition data of all robots in simulation during training. By enforcing the connectivity constraint and providing expert demonstrations, the policy learns to generate navigation actions with connectivity concerns. We compare different combinations of our key designs and validate the effectiveness of our approach. We also evaluate the generalization performance of our approach in both various simulated scenarios and real-world experiments.
As a case study of RL-based approach, this work presents an effective way to reduce exploration complexity with concerns of constraints satisfaction.


\bibliography{paper}

\begin{thebibliography}{10}
\providecommand{\url}[1]{#1}
\csname url@samestyle\endcsname
\providecommand{\newblock}{\relax}
\providecommand{\bibinfo}[2]{#2}
\providecommand{\BIBentrySTDinterwordspacing}{\spaceskip=0pt\relax}
\providecommand{\BIBentryALTinterwordstretchfactor}{4}
\providecommand{\BIBentryALTinterwordspacing}{\spaceskip=\fontdimen2\font plus
\BIBentryALTinterwordstretchfactor\fontdimen3\font minus
  \fontdimen4\font\relax}
\providecommand{\BIBforeignlanguage}[2]{{%
\expandafter\ifx\csname l@#1\endcsname\relax
\typeout{** WARNING: IEEEtran.bst: No hyphenation pattern has been}%
\typeout{** loaded for the language `#1'. Using the pattern for}%
\typeout{** the default language instead.}%
\else
\language=\csname l@#1\endcsname
\fi
#2}}
\providecommand{\BIBdecl}{\relax}
\BIBdecl

\bibitem{li2013bounded}
X.~Li, D.~Sun, and J.~Yang, ``A bounded controller for multirobot navigation
  while maintaining network connectivity in the presence of obstacles,''
  \emph{Automatica}, vol.~49, no.~1, pp. 285--292, 2013.

\bibitem{sabattini2013decentralized}
L.~Sabattini, N.~Chopra, and C.~Secchi, ``Decentralized connectivity
  maintenance for cooperative control of mobile robotic systems,'' \emph{The
  International Journal of Robotics Research (IJRR)}, vol.~32, no.~12, pp.
  1411--1423, 2013.

\bibitem{williams2015global}
R.~K. Williams, A.~Gasparri, G.~S. Sukhatme, and G.~Ulivi, ``Global
  connectivity control for spatially interacting multi-robot systems with
  unicycle kinematics,'' in \emph{2015 IEEE International Conference on
  Robotics and Automation (ICRA)}.\hskip 1em plus 0.5em minus 0.4em\relax IEEE,
  2015, pp. 1255--1261.

\bibitem{gasparri2017bounded}
A.~Gasparri, L.~Sabattini, and G.~Ulivi, ``Bounded control law for global
  connectivity maintenance in cooperative multirobot systems,'' \emph{IEEE
  Transactions on Robotics (T-RO)}, vol.~33, no.~3, pp. 700--717, 2017.

\bibitem{fiedler1973algebraic}
M.~Fiedler, ``Algebraic connectivity of graphs,'' \emph{Czechoslovak
  Mathematical Journal}, vol.~23, no.~2, pp. 298--305, 1973.

\bibitem{wang2016multi}
L.~Wang, A.~D. Ames, and M.~Egerstedt, ``Multi-objective compositions for
  collision-free connectivity maintenance in teams of mobile robots,'' in
  \emph{2016 IEEE 55th Conference on Decision and Control (CDC)}.\hskip 1em
  plus 0.5em minus 0.4em\relax IEEE, 2016, pp. 2659--2664.

\bibitem{capelli2020connectivity}
B.~Capelli and L.~Sabattini, ``Connectivity maintenance: Global and optimized
  approach through control barrier functions,'' in \emph{2020 IEEE
  International Conference on Robotics and Automation (ICRA)}.\hskip 1em plus
  0.5em minus 0.4em\relax IEEE, 2020, pp. 5590--5596.

\bibitem{pfeiffer2018reinforced}
M.~Pfeiffer, S.~Shukla, M.~Turchetta, C.~Cadena, A.~Krause, R.~Siegwart, and
  J.~Nieto, ``Reinforced imitation: Sample efficient deep reinforcement
  learning for mapless navigation by leveraging prior demonstrations,''
  \emph{IEEE Robotics and Automation Letters (RA-L)}, vol.~3, no.~4, pp.
  4423--4430, 2018.

\bibitem{long2018towards}
P.~Long, T.~Fan, X.~Liao, W.~Liu, H.~Zhang, and J.~Pan, ``Towards optimally
  decentralized multi-robot collision avoidance via deep reinforcement
  learning,'' in \emph{2018 IEEE International Conference on Robotics and
  Automation (ICRA)}.\hskip 1em plus 0.5em minus 0.4em\relax IEEE, 2018, pp.
  6252--6259.

\bibitem{lin2019end}
J.~Lin, X.~Yang, P.~Zheng, and H.~Cheng, ``End-to-end decentralized multi-robot
  navigation in unknown complex environments via deep reinforcement learning,''
  in \emph{2019 IEEE International Conference on Mechatronics and Automation
  (ICMA)}.\hskip 1em plus 0.5em minus 0.4em\relax IEEE, 2019, pp. 2493--2500.

\bibitem{lin2020connectivity}
{J. Lin, X. Yang, P. Zheng, and H. Cheng}, ``Connectivity guaranteed
  multi-robot navigation via deep reinforcement learning,'' in \emph{Conference
  on Robot Learning (CoRL)}.\hskip 1em plus 0.5em minus 0.4em\relax PMLR, 2020,
  pp. 661--670.

\bibitem{cheng2019end}
R.~Cheng, G.~Orosz, R.~M. Murray, and J.~W. Burdick, ``End-to-end safe
  reinforcement learning through barrier functions for safety-critical
  continuous control tasks,'' in \emph{Proceedings of the AAAI Conference on
  Artificial Intelligence}, vol.~33, no.~01, 2019, pp. 3387--3395.

\bibitem{mysore2020regularizing}
S.~Mysore, B.~Mabsout, R.~Mancuso, and K.~Saenko, ``Regularizing action
  policies for smooth control with reinforcement learning,'' in \emph{2021 IEEE
  International Confernece on Robotics and Automation (ICRA)}.\hskip 1em plus
  0.5em minus 0.4em\relax IEEE, 2021.

\bibitem{lowe2017multi}
R.~Lowe, Y.~Wu, A.~Tamar, J.~Harb, P.~Abbeel, and I.~Mordatch, ``Multi-agent
  actor-critic for mixed cooperative-competitive environments,'' \emph{Neural
  Information Processing Systems (NIPS)}, 2017.

\bibitem{foerster2016learning}
J.~N. Foerster, Y.~M. Assael, N.~de~Freitas, and S.~Whiteson, ``Learning to
  communicate with deep multi-agent reinforcement learning,'' in \emph{Neural
  Information Processing Systems (NIPS)}, 2016.

\bibitem{pfeiffer2017perception}
M.~Pfeiffer, M.~Schaeuble, J.~Nieto, R.~Siegwart, and C.~Cadena, ``From
  perception to decision: A data-driven approach to end-to-end motion planning
  for autonomous ground robots,'' in \emph{2017 IEEE International Conference
  on Robotics and Automation (ICRA)}.\hskip 1em plus 0.5em minus 0.4em\relax
  IEEE, 2017, pp. 1527--1533.

\bibitem{zhang2020map}
W.~Zhang, Y.~Zhang, and N.~Liu, ``Map-less navigation: A single drl-based
  controller for robots with varied dimensions,'' \emph{arXiv preprint
  arXiv:2002.06320}, 2020.

\bibitem{chen2017decentralized}
Y.~F. Chen, M.~Liu, M.~Everett, and J.~P. How, ``Decentralized
  non-communicating multiagent collision avoidance with deep reinforcement
  learning,'' in \emph{2017 IEEE International Conference on Robotics and
  Automation (ICRA)}.\hskip 1em plus 0.5em minus 0.4em\relax IEEE, 2017, pp.
  285--292.

\bibitem{everett2018motion}
M.~Everett, Y.~F. Chen, and J.~P. How, ``Motion planning among dynamic,
  decision-making agents with deep reinforcement learning,'' in \emph{2018
  IEEE/RSJ International Conference on Intelligent Robots and Systems
  (IROS)}.\hskip 1em plus 0.5em minus 0.4em\relax IEEE, 2018, pp. 3052--3059.

\bibitem{gupta2017cooperative}
J.~K. Gupta, M.~Egorov, and M.~Kochenderfer, ``Cooperative multi-agent control
  using deep reinforcement learning,'' in \emph{International Conference on
  Autonomous Agents and Multiagent Systems (AAMAS)}.\hskip 1em plus 0.5em minus
  0.4em\relax Springer, 2017, pp. 66--83.

\bibitem{nguyen2020deep}
T.~T. Nguyen, N.~D. Nguyen, and S.~Nahavandi, ``Deep reinforcement learning for
  multiagent systems: A review of challenges, solutions, and applications,''
  \emph{IEEE Transactions on Cybernetics}, vol.~50, no.~9, pp. 3826--3839,
  2020.

\bibitem{han2020cooperative}
R.~Han, S.~Chen, and Q.~Hao, ``Cooperative multi-robot navigation in dynamic
  environment with deep reinforcement learning,'' in \emph{2020 IEEE
  International Conference on Robotics and Automation (ICRA)}.\hskip 1em plus
  0.5em minus 0.4em\relax IEEE, 2020, pp. 448--454.

\bibitem{achiam2017constrained}
J.~Achiam, D.~Held, A.~Tamar, and P.~Abbeel, ``Constrained policy
  optimization,'' in \emph{International Conference on Machine Learning
  (ICML)}.\hskip 1em plus 0.5em minus 0.4em\relax PMLR, 2017, pp. 22--31.

\bibitem{schulman2015trust}
J.~Schulman, S.~Levine, P.~Abbeel, M.~Jordan, and P.~Moritz, ``Trust region
  policy optimization,'' in \emph{International Conference on Machine Learning
  (ICML)}.\hskip 1em plus 0.5em minus 0.4em\relax PMLR, 2015, pp. 1889--1897.

\bibitem{vecerik2017leveraging}
M.~Vecerik, T.~Hester, J.~Scholz, F.~Wang, O.~Pietquin, B.~Piot, N.~Heess,
  T.~Roth{\"o}rl, T.~Lampe, and M.~Riedmiller, ``Leveraging demonstrations for
  deep reinforcement learning on robotics problems with sparse rewards,''
  \emph{arXiv preprint arXiv:1707.08817}, 2017.

\bibitem{rajeswaran2017learning}
A.~Rajeswaran, V.~Kumar, A.~Gupta, G.~Vezzani, J.~Schulman, E.~Todorov, and
  S.~Levine, ``Learning complex dexterous manipulation with deep reinforcement
  learning and demonstrations,'' in \emph{Proceedings of Robotics: Science and
  Systems (RSS)}, 2018.

\bibitem{haarnoja2018soft}
T.~Haarnoja, A.~Zhou, K.~Hartikainen, G.~Tucker, S.~Ha, J.~Tan, V.~Kumar,
  H.~Zhu, A.~Gupta, P.~Abbeel \emph{et~al.}, ``Soft actor-critic algorithms and
  applications,'' \emph{arXiv preprint arXiv:1812.05905}, 2018.

\bibitem{engstrom2019implementation}
L.~Engstrom, A.~Ilyas, S.~Santurkar, D.~Tsipras, F.~Janoos, L.~Rudolph, and
  A.~Madry, ``Implementation matters in deep rl: A case study on ppo and
  trpo,'' in \emph{International Conference on Learning Representations
  (ICLR)}, 2019.

\bibitem{1606.01540}
G.~Brockman, V.~Cheung, L.~Pettersson, J.~Schneider, J.~Schulman, J.~Tang, and
  W.~Zaremba, ``Openai gym,'' \emph{arXiv preprint arXiv:1606.01540}, 2016.

\bibitem{tobin2017domain}
J.~Tobin, R.~Fong, A.~Ray, J.~Schneider, W.~Zaremba, and P.~Abbeel, ``Domain
  randomization for transferring deep neural networks from simulation to the
  real world,'' in \emph{2017 IEEE/RSJ International Conference on Intelligent
  Robots and Systems (IROS)}.\hskip 1em plus 0.5em minus 0.4em\relax IEEE,
  2017, pp. 23--30.

\bibitem{liu1989limited}
D.~C. Liu and J.~Nocedal, ``On the limited memory bfgs method for large scale
  optimization,'' \emph{Mathematical programming}, vol.~45, no.~1, pp.
  503--528, 1989.

\end{thebibliography}

\end{document}